\documentclass[conference]{IEEEtran}
\IEEEoverridecommandlockouts
\usepackage{cite}
\usepackage{amsmath,amssymb,amsfonts}
\usepackage{algorithm}
\usepackage{algorithmic}[1]
\usepackage{graphicx}
\usepackage{textcomp}
\usepackage{xcolor}
\newtheorem{theorem}{Theorem}
\usepackage{booktabs}
\usepackage{multirow}
\usepackage{bm}

\newtheorem{prop}{Proposition}

\newenvironment{proof}{\quad \textit{Proof:}}{\hfill$\square$}

\def\BibTeX{{\rm B\kern-.05em{\sc i\kern-.025em b}\kern-.08em
    T\kern-.1667em\lower.7ex\hbox{E}\kern-.125emX}}
\begin{document}

% \title{Immediate Learning on Large-Scale Streams: A Label-Decomposition-Based Online Forecasting Framework}
% \title{Immediate Learning on Large-Scale Streams: An Online Forecasting Framework for Streaming Data based on Label Decomposition}
\title{Act Now: A Novel Online Forecasting \\ Framework for Large-Scale Streaming Data}

% \thanks{This work was supported in part by the Joint Funds of the National Natural Science Foundation of China under Grant No. U22A2003 and U23A20277; National Natural Science Foundation of China under Grant No. 62271288; Mobility Programme of Sino-German Cooperation and Exchange NSFC under Grant No. M-0626.}

\author{\IEEEauthorblockN{Daojun Liang$^{1,5}$, Haixia Zhang$^{2,5}$, Jing Wang$^{3,5}$, Dongfeng Yuan$^{4,5}$, Minggao Zhang$^{2,5}$}
\IEEEauthorblockA{$^{1}$School of Information Science and Engineering, Shandong University, China\\
$^{2}$School of Control Science and Engineering, Shandong University, China\\
$^{3}$Ocean College, Jiangsu University of Science and Technology, China\\
$^{4}$School of Qilu Transportation, Shandong University, China\\
$^{5}$Shandong Key Laboratory of Intelligent Communication and Sensing-Computing Integration, China\\
liangdaojun@mail.sdu.edu.cn, haixia.zhang@sdu.edu.cn, jing.wang@just.edu.cn, \{dfyuan, mgzhang\}@sdu.edu.cn
}}

\maketitle
% \renewcommand{\thefootnote}{\fnsymbol{footnote}}
% \footnotetext[1]{Corresponding authors.}

\begin{abstract}
    In this paper, we find that existing online forecasting methods have the following issues: 1) They do not consider the update frequency of streaming data and directly use labels (future signals) to update the model, leading to information leakage.
    2) Eliminating information leakage can exacerbate concept drift and online parameter updates can damage prediction accuracy.
    3) Leaving out a validation set cuts off the model's continued learning.
    4) Existing GPU devices cannot support online learning of large-scale streaming data.
    To address the above issues, we propose a novel online learning framework, Act-Now, to improve the online prediction on large-scale streaming data. Firstly, we introduce a Random Subgraph Sampling (RSS) algorithm designed to enable efficient model training. 
    Then, we design a Fast Stream Buffer (FSB) and a Slow Stream Buffer (SSB) to update the model online. FSB updates the model immediately with the consistent pseudo- and partial labels to avoid information leakage. SSB updates the model in parallel using complete labels from earlier times. 
    Further, to address concept drift, we propose a Label Decomposition model (Lade) with statistical and normalization flows. Lade forecasts both the statistical variations and the normalized future values of the data, integrating them through a combiner to produce the final predictions. 
    Finally, we propose to perform online updates on the validation set to ensure the consistency of model learning on streaming data. 
    Extensive experiments demonstrate that the proposed Act-Now framework performs well on large-scale streaming data, with an average 28.4\% and 19.5\% performance improvement, respectively.
    Experiments can be reproduced via https://github.com/Anoise/Act-Now.
\end{abstract}


\section{Introduction}
Streaming data have become an essential part of our daily lives, and the demand of streaming services is growing rapidly.
The exponential growth of streaming data poses significant challenges, including the serious energy consumption \cite{narayanan2021variegated} and the inability to meet differentiated quality of service requirements \cite{asghar2022evolution}.
Streaming data forecasting can guide manager to achieve long-term planning and decision-making \cite{niu2010cell} \cite{kato2016deep, DeepCog2019}, which is an important means to achieve green and intelligent applications \cite{li2017intelligent}. 
Currently, streaming data forecasting is studied as online time series prediction and has received widespread attention.
It is desirable to train a forecaster online \cite{anava2013online,liu2016online} using only new samples to capture the changing dynamic in the environment.

\begin{figure}%[htbp]
    \centerline{\includegraphics[width=0.9\columnwidth]{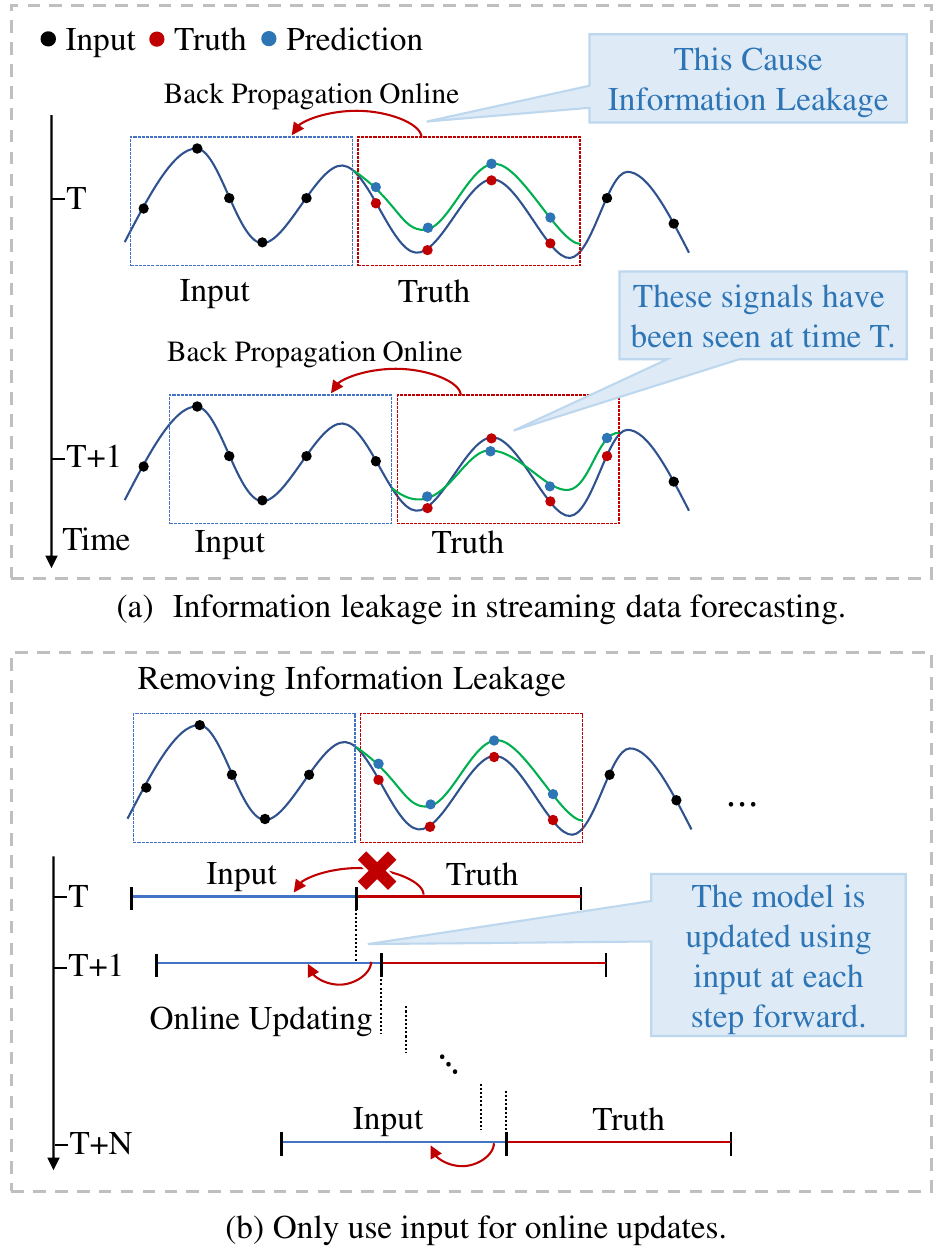}}
    \caption{(a) Information leakage in streaming data forecasting. (b) Removing information leakage requires using the input to update the model online. 
    }
    \label{fig_info_leak} % A lower MSE indicates a better performance.
\end{figure}

\begin{figure*}%[htbp]
    \centerline{\includegraphics[width=\textwidth]{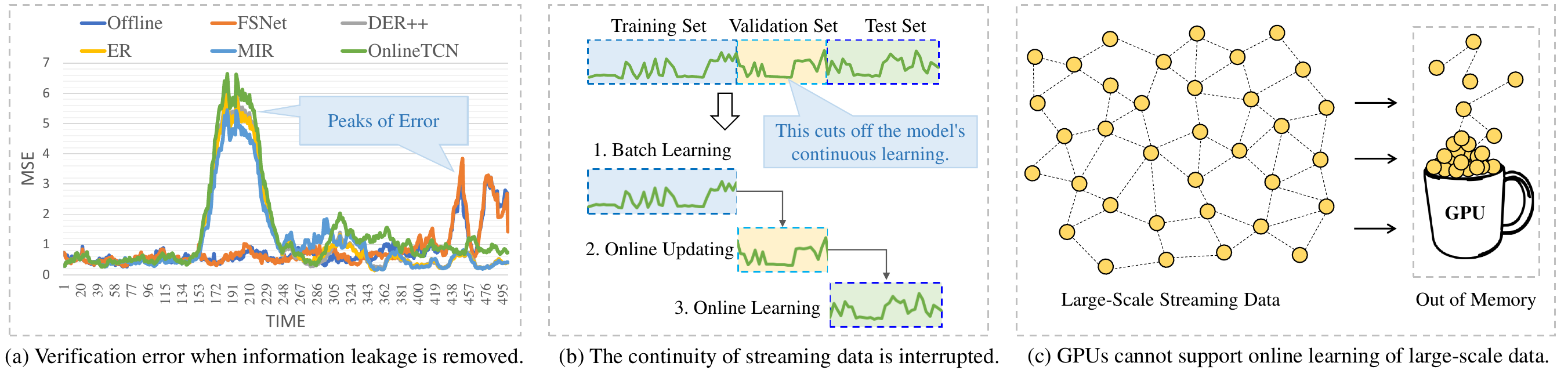}}
    \caption{(a) Removing information leakage exacerbate concept drift, and model updating online may damage forecasting. (b) Leaving out the validation set alone cuts off the model's continued learning. (c) Existing GPU devices cannot support online learning of large-scale streaming data.
    }
    \label{fig_issues} % A lower MSE indicates a better performance.
\end{figure*}

Although some studies \cite{smola2003laplace, liu2016online, gultekin2018online, kurle2019continual, aydore2019dynamic,  woo2022cost, zhang2023onenet, pham2023learning} have started to tackle the concept drift of long-term forecasting through online learning, they overlook several critical issues in online forecasting:

Firstly, existing methods fail to account for the update frequency of streaming data, i.e., the timing of streaming data arrivals is not considered. As shown in Fig. \ref{fig_info_leak}(a), they directly use labels (which contain unseen signals) to update the model, resulting in information leakage.

Secondly, when information leakage is removed, the model lacks sufficient supervised data to guide its updates, relying solely on the most recent input data available at the next time step, as illustrated in Fig. \ref{fig_info_leak}(b). This prevents the model from being updated immediately, further exacerbating concept drift and impairing the performance of forecasting. 
As shown in Fig. \ref{fig_issues}(a), existing online learning methods, when lacking sufficient supervisory signals to update the model, tend to focus excessively on the present while losing forward-looking capabilities. This further results in rapid performance degradation, significant error peaks, and potentially inferior overall performance compared to offline methods.

Third, the current online learning framework divides the streaming data into training, validation, and test sets: batch training is performed on the training set, the model with the smallest generalization error is selected using the validation set, and online predictions are then performed on the test set, as shown in Fig. \ref{fig_issues}(b). The problem with this approach is that the validation set is separated from the streaming data, breaking the continuity of the streaming data, making it learn earlier and test later. This violates the assumption that online learning solves concept drift but further aggravates the issue.

Lastly and importantly, in real-world applications, sensor nodes are densely distributed and present a graph-like structure, making it impossible to efficiently perform long-term forecasting tasks. 
For example, there are more than 20,000 nodes in a medium-sized city, and each of them requires long-term time series forecasting, which will result in a single GPU being unable to handle such large-scale streaming data. 

To address the above issues, we propose a novel online learning framework (Act-Now), which improves the online forecasting framework in several aspects:

Firstly, we developed a Random Subgraph Sampling (RSS) algorithm, which, according to our findings, effectively addresses the training challenges of deep models on large-scale irregular streaming data.
The proposed RSS algorithm enables the disassembly of large-scale streaming data into multiple small-scale subseries through random subgraphs sampling, allowing each to be processed on existing GPU devices.

Then, we considered the continuity of streaming data and adopted an incremental storage approach, saving data only once for each time step. Based on this, we designed a Fast Stream Buffer (FSB) and a Slow Stream Buffer (SSB) to update the model online. 
FSB updates the model immediately by utilizing the consistency loss of partial labels to prevent performance degradation resulting from delayed updates. SSB uses the earlier complete labels to update the model in parallel on other GPUs or devices. 

Further, to address concept drift, we propose a Label Decomposition model with statistical and normalization flows, namely Lade. Lade utilize decomposition to deconstruct a time series into several more predictable components and leverage deep models to learn the time-varying patterns of each component separately.
Instead of treating decomposition as a preprocessing step \cite{taylor2018forecasting, oreshkin2019n, sen2019think} or embedding it in a learning module \cite{wu2021autoformer, zhou2022fedformer, liu2022non}, Lade uses decomposition for supervised signals. Meanwhile, Lade forecasts both the statistical variations and the normalized future values of the data, integrating them through a combiner to produce the final predictions.

In addition, we propose to perform online updates on the validation set to ensure the consistency of model learning on streaming data. This ensures that the model learns on a continuous data stream, which is more conducive to mitigating the issue of concept drift.

In summary, we propose a novel online learning framework, namely Act-Now, that differs significantly from traditional online learning methods. It includes large-scale streaming data processing, the design of streaming buffers, a consistency loss that enables immediate model updates, and the effective utilization of the validation set.
Extensive experimental results show the proposed method outperforms existing methods by a large margin, average 28.4\% and 19.5\% performance improvement on the large-scale streaming datasets. 
In addition, we build an open source library of Act-Now to facilitate related research, which  provides a strong benchmark at https://github.com/Anoise/Act-Now.

\begin{figure*}
  \centerline{\includegraphics[width=\textwidth]{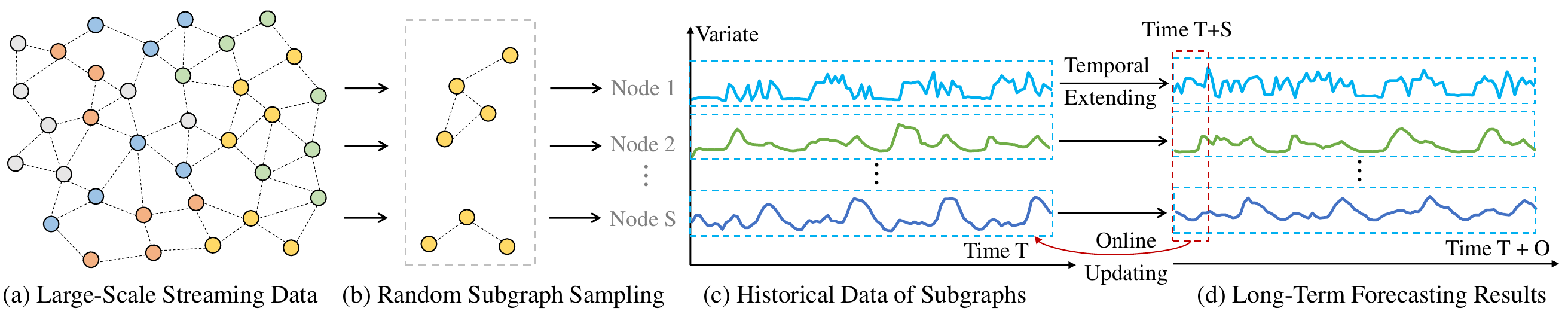}}
  \caption{Random Subgraph Sampling (RSS): For large-scale graph-structured data (a), a subgraph is randomly selected at each iteration (b). Through multiple sampling iterations, comprehensive coverage of the large-scale graph is achieved, enabling the full utilization of its node attributes and structural information. During the network training phase (repeated) and testing phase (non-repeated), long-term wireless traffic prediction (d) is performed by temporally extending the historical data of the subgraph (c).}
  \label{fig_scene}
\end{figure*}

The remainder of the paper is organized as follows:
Firstly, related works are presented in Section \ref{sec_rework}.
In Section \ref{sec_rss}, we first introduce the RSS method.
Then, in Section \ref{sec_buffer}, the FSB and SSB mechanisms are presented.
Next, the architecture of Lade in Section \ref{sec_lade} is introduced. 
Section \ref{sec_exp} presents the experimental results of the proposed method. 
Finally, we conclude this paper in Section \ref{sec_con}.

\section{Related Work}
\label{sec_rework}

%%% 重新写；

\subsection{Concept Drift} 

Real-world concepts are often dynamic and subject to change over time, a characteristic particularly evident in domains such as customer preference analysis and traffic forecasting. Due to shifts in the underlying data distribution, models trained on historical data may become misaligned with new observations, necessitating frequent updates to preserve their accuracy. This phenomenon, referred to as concept drift \cite{tsymbal2004problem}, introduces additional complexity in the model learning process. In this paper, we address the challenge of online learning for time series forecasting \cite{li2022ddg,qin2022generalizing,pham2023learning}.

\subsection{Online Time Series Forecasting}

Online time series forecasting \cite{liu2016online,gultekin2018online,aydore2019dynamic} is prevalent in numerous real-world applications due to the inherently sequential nature of the data. In this paradigm, there is no distinct separation between training and evaluation. Instead, learning occurs iteratively over a series of rounds. In each round, the model is provided with a look-back window and tasked with predicting the forecast window. Afterward, the true outcome is revealed, which serves to enhance the model's predictions for subsequent rounds. The model's performance is typically assessed based on its cumulative prediction errors over time \cite{sahoo2018online}. Given the inherent challenges of this setting, online time series forecasting entails several sub-problems, ranging from learning under concept drift  \cite{gama2014survey} to handling missing values resulting from irregularly sampled data. This research focuses on the issue of fast learning, specifically in terms of sample efficiency, under conditions of concept drift, by improving the architecture of deep networks and leveraging relevant past knowledge. 
While a substantial body of literature on Bayesian continual learning exists for addressing regression tasks \cite{smola2003laplace,kurle2019continual}, such approaches typically adhere to the Bayesian framework, which permits the forgetting of prior knowledge and lacks a mechanism for rapid learning \cite{huszar2017quadratic,kirkpatrick2018reply}. Furthermore, these methods have not been applied to deep neural networks, and adapting them to the context of this study presents significant challenges.

\subsection{Continual Learning}

Continual learning \cite{lopez2017gradient} is an emerging area of research focused on developing intelligent agents capable of learning to perform a sequence of tasks over time, while having only limited access to past experiences. A continual learner must strike an effective balance between retaining the knowledge acquired from previous tasks and facilitating the learning of new tasks, a challenge known as the stability-plasticity dilemma \cite{grossberg1982does}. Motivated by parallels with human learning, several neuroscience-based frameworks have inspired the development of various continual learning algorithms. One widely recognized framework is the Complementary Learning Systems (CLS) theory, which posits a dual learning system. Continual learning approaches derived from CLS theory enhance slow, deep networks with the capacity to rapidly learn from data streams \cite{lin1992self,riemer2018learning,rolnick2019experience}, either through experience replay mechanisms or by explicitly modeling the fast and slow learning components \cite{pham2021dualnet,arani2021learning}. These methods have shown promising results on controlled benchmarks in vision and language. In contrast, our work tackles the challenges of online time series forecasting by framing them as a continual learning problem.

\begin{figure*}
    \centerline{\includegraphics[width=0.98\textwidth]{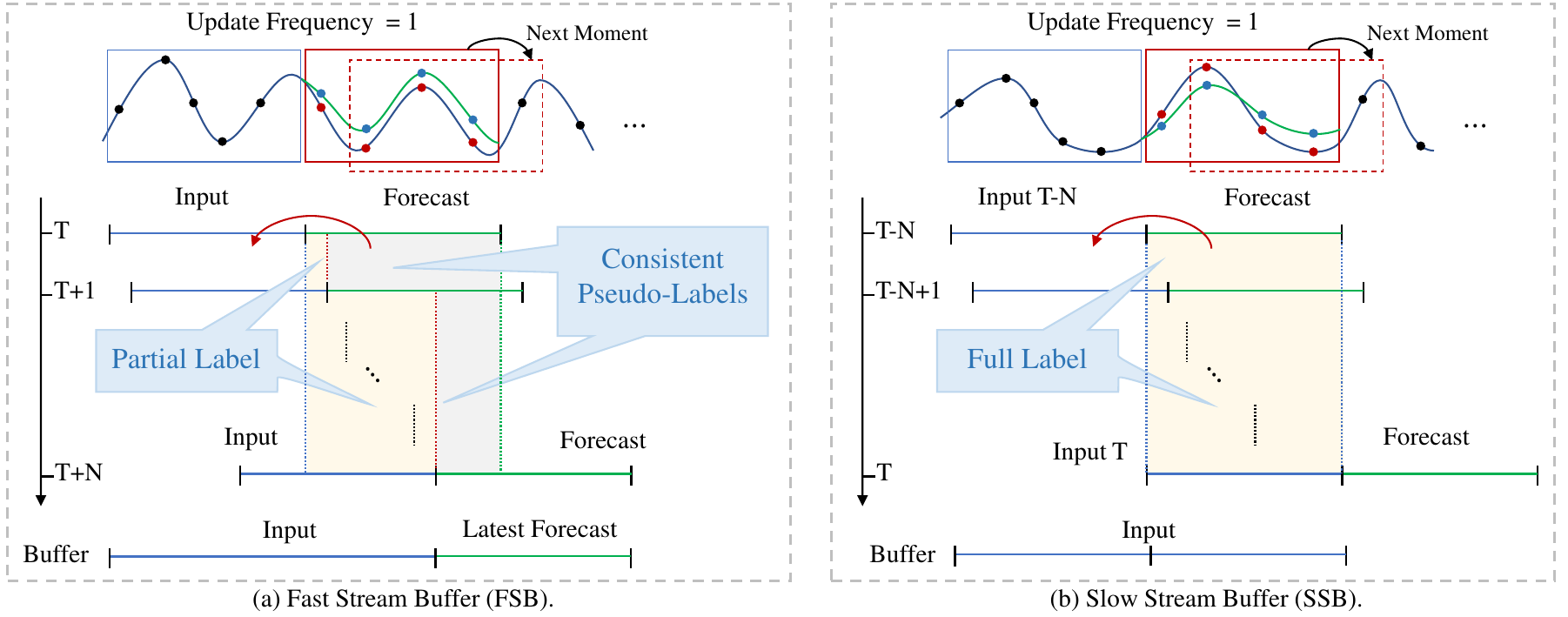}}
    \caption{Both FSB and SSB are implemented through streaming buffers (bottom of the figures). (a) FSB updates the model online through partial labels and consistent pseudo labels, while SSB (b) updates the model online through full labels at an earlier time.
    }
    \label{fig_buffer}
\end{figure*}

\section{Random Subgraph Sampling (RSS)}
\label{sec_rss}

Random Subgraph Sampling (RSS) aims to reduce the computational burden of training deep models on large graphs while maintaining the model's generalization performance. RSS randomly selects a subgraph $G_{sub}$ from the original large-scale graph $G=(V,E)$ to train the model $\mathcal{F}$, where $V$ indicates nodes and $E$ indicates edges.
As shown in Fig. \ref{fig_scene}.a, the original large-scale graph is disassembled into multiple small subgraphs $\{X_{sub}, Y_{sub}, E_{sub}\}$ through RSS, so that each of them can be processed on existing GPU devices.
The following theorem proves that RSS offers comprehensive coverage of the original large-scale graph, which is an unbiased estimator of the true aggregated feature using the entire graph.
\begin{theorem}
  \textit{With sufficient sampling, the subgraph generated by RSS sampling serves as an unbiased estimator, leveraging the true aggregated features of the entire graph.}
  \label{th1}
\end{theorem}

The proof of Theorem \ref{th1} is given in Appendix \ref{ssec_prf_th1}.
Theorem \ref{th1} means that we can train a well-performing learning model by randomly sampling subgraphs from a large graph.
We provide a Python pseudocode implementation of RSS in Algorithm~\ref{alg_rss} of Appendix \ref{ssec_rss_alg}. RSS facilitates random sampling of large graphs across multiple batches, thereby enabling efficient training of deep models.

\section{FSB and SSB}
\label{sec_buffer}

%%% 重新写
\subsection{Existing Online Forecasting Framework} 

The purpose of streaming data forecasting is to use the observed value of $L_{in}$ historical moments to predict the missing value of $L_{out}$ future moments, which can be denoted as $Input \text{-}L_{in}\text{-}predict\text{-}L_{out}$. If the feature dimension of the series is denoted as $D$, its input data at time $t$ can be denoted as $X^t = \{s_1^t, \cdots, s_{L_{in}}^t | s_k^t \in \mathbb{R}^D  \}$, and its output at time $t$ can be denoted as $Y^t = \{s_{L_{in}+1}^t, \cdots, s_{L_{in}+L_{out}}^t | s_{L_{in}+k}^t \in \mathbb{R}^D  \}$. 
Then, we can predict $Y^t \in \mathbb{R}^{L_{out}\times D}$ by designing a model $\mathcal{F}$ given an input $X^t \in \mathbb{R}^{L_{in}\times D}$, which can be expressed as: $Y^t = \mathcal{F}(X^t)$. 
For denotation simplicity, the superscript $t$ will be omitted if it does not cause ambiguity in the context.

In real-world applications, the model builds on the historical data needs to forecast the future data, that is, given time offset $O>L_{in}$, and $X_{O-L_{in}+1:O}$, the model needs to forecast $Y_{O+1:O+L_{out}}$. 
Online streaming data forecasting is a widely adopted technique in real-world applications, driven by the sequential nature of data and the frequent occurrence of concept drift. In this framework, learning unfolds over a series of rounds: the model is provided with a look-back window to generate predictions for a forecast window. Subsequently, the true values are revealed, which is used to refine the performance of the model in subsequent rounds. During online adaptation, the model is updated using the incoming data stream, optimized with the Mean Squared Error (MSE) loss calculated for each channel:
\begin{equation}
    \mathcal{L}=\frac{1}{T}\sum_{t=1}^T \parallel Y^t_{O+1:O+L_{out}} - \hat{Y}^t_{O+1:O+L_{out}} \parallel. 
\end{equation}

\subsection{Update Frequency of Streaming Data}

Previous online forecasting framework did not consider the update frequency $D_{freq}$ of streaming data. They assumed an update frequency of 1 and used long-term full labels $Y^t_{O+1:O+L_{out}}$ to update the data, leading to information leakage.
However, at the current moment $t$, only $X_{O-L_{in}+1:O}$ can be exploited for online learning.
The update frequency is crucial for online forecasting methods, as it determines the availability of labels over time and the timing of model updates. Consequently, it directly impacts the model's predictive performance and its evaluation methods.
To better represent and model the update frequency while removing information leakage, we propose a streaming buffer mechanism.

\subsection{Streaming Buffer Mechanism}

Considering the continuity of streaming data, we propose an incremental buffer mechanism that saves data only once at each time step. 
When retrieving data from the buffer, we use a sliding window to extract the inputs and true values according to the update frequency.
Unlike previous buffer mechanisms that stored large amounts of redundant data at each time step, the incremental buffer mechanism retains only data from distinct time points, greatly reducing memory overhead.
Based on this, we designed a Fast Stream Buffer (FSB) and a Slow Stream Buffer (SSB) to update the model online. 
FSB updates the model online using partial labels and consistent pseudo labels, while SSB updates the model in parallel using full labels at an earlier time.

\subsection{Consistent Pseudo-Labels}

As shown in Fig. \ref{fig_buffer}(a), FSB ensures the model remains responsive to real-time data by quickly incorporating updates based on partial labels and consistent pseudo-labels, thus reducing the risk of performance degradation due to delayed learning. 
Specifically, when we only have access to partial labels $Y_{part}$, we can make use of pseudo-labels:
\begin{prop}
 (\textbf{Consistent Pseudo-Labels}) \textit{For predictions over a future time period, assume that results $\hat{Y}_{new}$ derived from recent inputs are superior to those $\hat{Y}_{old}$ based on inputs from more distant time points. 
 Therefore, results $\hat{Y}_{new}$ can be used as consistent pseudo-labels for forecasts $\hat{Y}_{old}$.
 } 
\label{prop1}
\end{prop}

Consistent pseudo-labels play a critical role in enhancing the effectiveness of FSB by providing reliable guidance for model updates when complete labels are unavailable. 
These pseudo-labels are generated based on the model's confidence and consistency across data points, ensuring they align closely with the true data distribution.
Specifically,
\begin{align}
  \mathcal{L}_{cpl} & = \text{Distance} \left( [Y_{part}, \hat{Y}_{new}] , \hat{Y}_{old}  \right) , \label{eq_cpl}
\end{align}
By incorporating consistent pseudo-labels into the update process, FSB can simulate the impact of fully labeled data, enabling more accurate and meaningful adjustments to the model in real time. 
The immediate update mechanism enables the model to adapt swiftly to changing data distributions, maintaining its predictive performance in dynamic environments. The SSB updates the model online using full labels from earlier time points and can perform updates on other devices to reduce the computational load on the local device.
This not only improves the model's ability to generalize but also mitigates the potential noise or instability in the progress of online learning.

\subsection{Parallel Online Updates}

SSB updates the model online using full labels from earlier time points. Therefore, it can be deployed on other GPUs or devices to reduce the computational load on the local device.
Here, we consider the scenario of updating the model $M$ in parallel on GPU $g$:
\begin{align}
    X^g_{input} & = \text{Loading}(\mathcal{B}^g_{ssb}, D_{freq}) \\
  \hat{Y} & = M^g_{copy}(X^g_{input}), \label{eq_parall}
\end{align}
where $\mathcal{B}^g_{ssb}$ is the SSB initialized on GPU $g$, $M^g_{copy}$ represents copying model $M$ to GPU $g$.
This parallelized approach not only improves computational efficiency but also ensures the model benefits from more robust training, integrating fully labeled data without interrupting real-time operations. 

% Together, the FSB and SSB enable a balance between responsiveness and thoroughness, ensuring the system can adapt quickly while leveraging the depth of complete data when available.

\subsection{Online Updates on the Validation Set}

As shown in Fig. \ref{fig_issues}(b), we propose to perform online updates on the validation set to ensure the consistency of model learning on streaming data. Online updates on the validation set serve a dual role in the context of streaming data: maintaining model consistency and addressing concept drift. 
By incorporating online updates, the validation set transitions from a static evaluation mechanism to an active participant in the learning process, ensuring that the model adapts dynamically to evolving data distributions. 
This approach is particularly advantageous in mitigating concept drift, as it enables the model to continuously align its predictions with the most recent data patterns. 
Furthermore, online updates enhance the representativeness of the validation set, ensuring that it accurately reflects the current state of the data stream.

\begin{figure*}
  \centerline{\includegraphics[width=\textwidth]{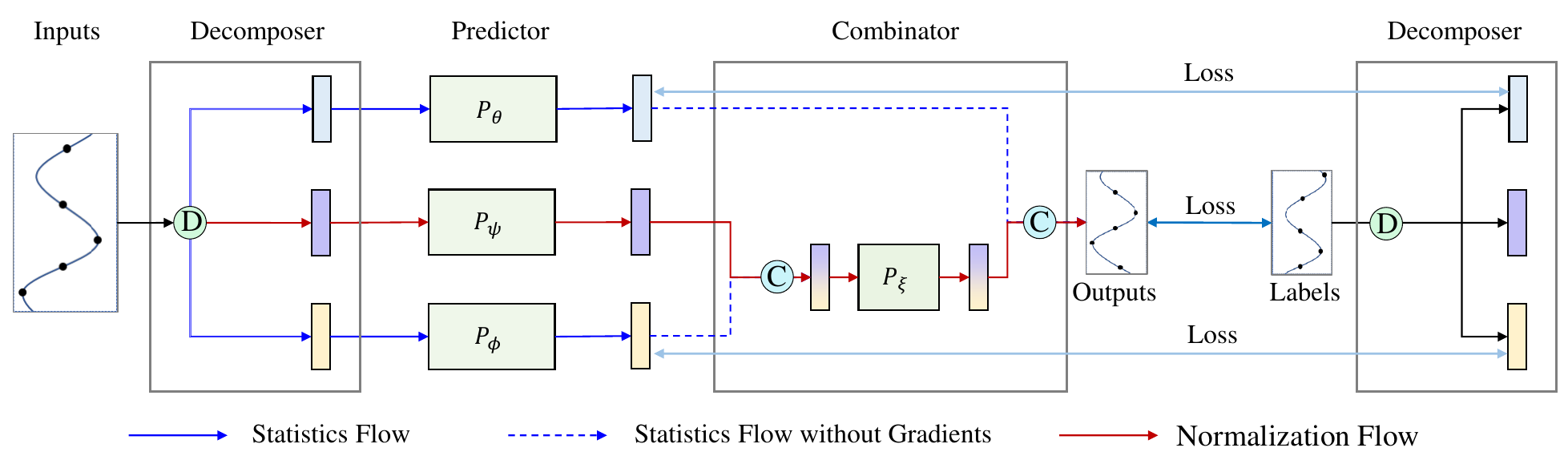}}
  \caption{The architecture of Lade, which including three main parts: decomposer $D$, predictor $P$, as well as combinator $C$. The decomposer is tasked with decomposing the input series into a statistical flow and a normalization flow. The predictors are employed to model the nonlinear components of the prediction, while the predictor aligns the outputs of the individual learners with the decomposed components of the label. The combinator integrates the predictions of each component to derive the final output. During the learning process, each label component is back-propagated to the shallow layers to gradually supervise their learning process.}
  \label{fig_arch} 
\end{figure*}

\section{Lade}
\label{sec_lade}

Decomposition is a powerful technique for analyzing and solving non-stationary issues in time series data \cite{cleveland1990stl, de2011forecasting, cheng2015time}. Decomposition addresses non-stationary by breaking the time series into distinct components that can be more easily analyzed and modeled. The primary components typically considered are mean, variance, trend, seasonal, and residual (or noise) components.
In this section, we introduce Lade, a method designed to mitigate concept drift in streaming data. As shown in Fig. \ref{fig_arch}, Lade decomposes a time series into a statistics flow and a normalization flow, utilizing deep models to independently learn the time-varying patterns of each.

\subsection{Statistical and Normalization Flow}

The statistical and normalization flow of Lade serves as a foundational mechanism for managing concept drift by enhancing the clarity and stability of data patterns over time. The statistical component ensures that the time series is effectively decomposed into more predictable subcomponents, each representing distinct and more stable patterns. This decomposition reduces the complexity and noise often associated with raw time series data, enabling more focused and efficient learning. 
The normalization flow further stabilizes the learning process by scaling the decomposed components into a consistent range, mitigating the influence of outliers or abrupt shifts. Together, these processes ensure that the deep models can concentrate on time-varying dynamics within each component, leading to improved adaptability to evolving data distributions. 
The combination of statistical and normalization flows enables Lade to effectively address concept drift, enhancing robustness and generalization across diverse and dynamic environments.
To achieve statistical and normalization flow, as shown in Fig. \ref{fig_arch}, the entire architecture consists of three main components: the decomposer, the predictor, and the combiner.

\subsection{Decomposer}

The decomposer is responsible for decomposing the target variable $Y$ into components that capture non-stationary and stationary behaviors respectively.
Many classic time series decomposition methods, including additive decomposition $Y = T + S + N$, multiplicative decomposition $Y = T\times S \times N$, and seasonal-trend decomposition $Y = T + S + N$, can be employed for label decomposition, 
where $T$ captures the long-term trend, $S$ captures the seasonal or cyclical patterns, and $N$ captures the residual stationary component.
In this paper, we utilize a mean-variance decomposition. 

\textbf{Mean-Variance Decomposer} (MVD): 
The mean-variance decomposer is a hybrid decomposition method that combines the advantages of both additive and multiplicative decompositions.
The decomposition of the target variable by MVD can be expressed as:
\begin{align}
  M & = \text{Mean}(Y, -1), \notag \\
  Y' & = Y - M, \notag \\
  V & = \text{Var}(Y', -1), \notag \\
  N & = Y' / V + \epsilon, \label{eq_mvd}
\end{align}
where $M$ and $V$ are the mean and variance of $Y$, respectively, $\epsilon$ is a small constant to avoid division by zero. 
The mean $M$ and variance $V$ in Eq. \ref{eq_mvd} are the statistical information of the statistical flow, while the variable $N$ in Eq. \ref{eq_mvd} corresponds to the normalization flow.

\subsection{Predictor}

To progressively guide the model's learning process, we employ different predictors to give predictions for each component individually at the model's shallow level.
To achieve this, we insert a decomposer at the input to decompose it into its component forms corresponding to those of the output, like
\begin{align}
  M_X & = \text{Mean}(X, -1), \notag \\
  X' & = X - M_X, \notag \\
  V_X & = \text{Var}(X', -1), \notag \\
  N_X & = X' / V_X + \epsilon, \label{eq_x_mvd}
\end{align}
Indeed, the predictor can be any learnable models.
Let $P$ be the predictor parameterized by $\theta$, $\phi$ and $\psi$, respectively. 
For MVD, we have:
\begin{align}
  \hat{M}, \hat{V},  \hat{N} = P_\theta(M_X)), P_\phi(V_X)), P_\psi(N_X)). \label{eq_learner}
\end{align}
Therefore, the statistical information $\hat{M}$ and $\hat{V}$, as well as the normalized flow $\hat{N}$, can be obtained separately through the predictors.
They are then combined to get the final predictions.

\subsection{Combinator}

The combiner integrates the predictions of each component to produce the final prediction results, which is the inverse of the decomposition process.
As shown in Fig. \ref{fig_arch}, 
The components are sequentially integrated in the combiner, which incorporates learnable parameters to coordinate the contributions of each component.
Specifically, for MVD, the prediction results of combiner can be expressed as:
\begin{align}
  \hat{Y} & = P_{\xi}(\hat{V}_{detach} \times \hat{N}) + \hat{M}_{detach}  , \label{eq_cbn_mvd}
\end{align}
where $\xi$ is the parameter of the combiner, $\hat{M}_{detach}$ and $\hat{V}_{detach}$ are the gradient-free versions of $\hat{M}$ and $\hat{V}$,  respectively.

\subsection{Loss and Optimization}

The loss function of Lade comprises two parts: the prediction loss of statistical flow generated by the predictor, and the final prediction loss of normalization flow produced by the combiner.
For MVD, the loss function of the statistical flow can be expressed as:
\begin{align}
  \mathcal{L}_{stat} & = \mathcal{L}_{\theta}(\hat{M}, M) + \mathcal{L}_{\phi}(\hat{V}, V) , \label{eq_loss_mvd}
\end{align}
and the loss function of the normalization flow can be expressed as:
\begin{align}
  \mathcal{L}_{norm} & = \mathcal{L}_{\xi, \psi}(\hat{Y}, Y) . \label{eq_loss_cbn}
\end{align}

We set up two optimizers to minimize $\mathcal{L}_{stat}$ and $\mathcal{L}_{norm}$ separately, instead of using a single optimizer to minimize the sum of them.
This approach is designed to clearly separate the statistical flow and the normalization flow, avoiding mutual confusion between them, which could result in learning irrelevant information and ultimately degrade model performance.

% \subsection{Accelerating Lade}

% \begin{figure}
%   \centerline{\includegraphics[width=\columnwidth]{figs/fast_Lade.pdf}}
%   \caption{The accelerated Lade architecture. The decomposed components, along with the learners and predictors, are combined into longer vectors or wider models. This parallelization can greatly accelerate the training and inference process.
%   }
%   \label{fig_fast_arch} %  A lower MSE indicates a better performance. 
% \end{figure}

% It is observed that the learner and predictor in Lade operate in parallel. If they share the same model structure, they can be merged to enhance computational efficiency.
% As shown in Fig. \ref{fig_fast_arch}, the learners and predictors are merged into a single model, and their parameters are combined into a single set $[\theta, \phi, \psi ]$. 
% This is equivalent to widening the original learner and predictor, which can be expressed as:
% \begin{align}
%   [\hat{M},\hat{V},\hat{R}] & = P_{[\theta, \phi, \psi ]}\left(L_{[\theta, \phi, \psi ]}([M_X, V_X, R_X]) \right) . \label{eq_fast_mvd}
% \end{align}
% Similarly, the component loss can be expressed as:
% \begin{align}
%   \mathcal{L}_{cpn} & = \mathcal{L}_{[\theta, \phi, \psi ]}([\hat{M},\hat{V},\hat{R}], [M, V, R]). \label{eq_fast_loss}
% \end{align}
% By employing Eq. \ref{eq_fast_loss}, we can train a single integrated model, which is typically more efficient than training multiple independent models due to the parallel operations.

\begin{table*}[!t]
    \centering
    \caption{Online learning results on three large-scale datasets. Backbone uses Lade if a method is model-independent.}
    \label{tb_on_main}
    \resizebox{0.9\textwidth}{!}
  {
    \begin{tabular}{cc| ccc| ccc| ccc}
    \toprule
    \multicolumn{2}{c}{Datasets}         & \multicolumn{3}{c}{CBS ($L_{in}=36, N_{part}=17$)}                     & \multicolumn{3}{c}{Milano ($L_{in}=36, N_{part}=25$)}     & \multicolumn{3}{c}{C2TM ($L_{in}=8, N_{part}=62$)} \\ \midrule
    Methods                    & Lengths & 24             & 48             & 72         & 24        & 48        & 72     & 4      & 6      & 8      \\ \midrule
    \multirow{2}{*}{Offline}   & MSE     & 3.054          & 3.458          & 3.911      & 2.330     & 2.607     & 3.725  & \bf 1.974  & 2.078  & 2.166  \\
                               & MAE     & 0.871          & 0.886          & 0.910      & 0.738     & 0.752     & 0.831  & 0.084  & 0.079  & 0.070  \\ \midrule
    \multirow{2}{*}{DER++}     & MSE     & 3.765          & 4.190          & 4.550      & 2.139     & 2.543     & 3.619  & 1.976  & 2.079  & 2.172  \\
                               & MAE     & 0.992          & 1.017          & 1.038      & 0.625     & 0.664     & 0.741  & 0.071  & 0.074  & 0.073  \\ \midrule
    \multirow{2}{*}{ER}        & MSE     & 2.979          & 3.392          & 3.834      & 1.988     & 2.473     & 3.382  & 1.976  & 2.080  & 2.175  \\
                               & MAE     & 0.854          & 0.871          & 0.895      & 0.593     & 0.642     & 0.693  & 0.072  & 0.077  & 0.079  \\ \midrule
    \multirow{2}{*}{FSNet}     & MSE     & 10.385         & 11.161         & 11.783     & 4.001     & 4.723     & 5.419  & 2.041  & 2.121  & 2.213  \\
                               & MAE     & 1.447          & 1.510          & 1.575      & 0.896     & 0.935     & 0.993  & 0.088  & 0.087  & 0.085  \\ \midrule
    \multirow{2}{*}{OnlineTCN} & MSE     & 3.031          & 3.435          & 3.884      & 2.322     & 3.190     & 3.725  & \bf 1.974  & 2.076  & 2.173  \\
                               & MAE     & 0.866          & 0.884          & 0.906      & 0.713     & 0.779     & 0.813  & 0.079  & 0.076  & 0.071  \\ \midrule
    \multirow{2}{*}{MIR}       & MSE     & 2.958          & 3.371          & 3.807      & 1.973     & 2.371     & 2.355  & 1.976  & 2.082  & 2.176  \\
                               & MAE     & 0.850          & 0.867          & 0.892      & 0.589     & 0.621     & 0.617  & 0.073  & 0.078  & 0.079  \\ \midrule
    \multirow{2}{*}{Lade}      & MSE     & \bf 2.547      & \bf 3.068      & \bf 3.490  & \bf 1.797 & \bf 2.196 & \bf 2.304  & 1.984  & \bf 2.075  & \bf 2.152  \\
                               & MAE     & \bf 0.723      & \bf 0.765      & \bf 0.794  & \bf 0.439 & \bf 0.481 & \bf 0.487  & \bf 0.062  & \bf 0.078  & \bf 0.070   \\ \bottomrule
    \end{tabular}
}
\end{table*}

\section{Experiments}
\label{sec_exp}

The proposed methods are extensively evaluated on the three widely used real-world large-scale datasets, including multiple mainstream large-scale streaming datasets. 

\subsection{Datasets}

1) Milano \cite{barlacchi2015multi} is provided by SKIL\footnote{http://jol.telecomitalia.com/jolskil} of Telecom Italia. The dataset was collected from November 1, 2013, to January 1, 2014, and the data is aggregated into 10-minute intervals over the whole city of Milan (62 days, 300 million records, about 19 GB).
Milano records the time of the interaction and the specific radio base station that managed it. 
The dataset contains 10,000 nodes, and the length of each node is 1498. The ratio of training, validation and test datasets is 10:2:3.

2) CBS is collected from all base station in a certain city in China, including up-link and
down-link communication trafﬁc data. The dataset is collected from June 8 to July 26, 2019, with a temporal interval of 60 minutes over the whole city (94 days, 51.9 million records, about 17.1 GB). 
CBS contains 4454 nodes, each with a length of 4032. The ratio of training, validation and test datasets is 3:1:1.

3) C2TM \cite{Modeling15Chen} makes an analysis on week-long traffic generated by a large population of people in a median-size city of China. 
C2TM analyses make use of request-response records extracted from traffic at the city scale, consisting of individuals' activities during a continuous week (actually eight days from Aug. 19 to Aug. 26, 2012), with accurate timestamp and location information indicated by connected cellular base stations. 
C2TM is a dataset characterized by shorter sequences and higher noise levels, used to evaluate the model's ability to handle highly volatile data.
This dataset contains 13269 nodes, each with a length of 192. The ratio of training, validation and test datasets is 5:1:2.

\subsection{Implementation Details}

For the large-scale datasets, we initially employ the RSS algorithm to perform data sampling. In this algorithm, we random decompose the entire large graph into 17, 25, 62 sub-graphs for the CBS, Milano and C2TM datasets, respectively.
The number of hidden units is set to 512. 
The model is trained using the ADAM \cite{kingma2014adam} optimizer and L2 loss.
The learning rate is set to 0.0001. 
The batch size is set to 64 for training and 1 for testing, and the maximum number of epochs is set to 10.
For the predictor in Lade, we employ a 2-layer MLP with ReLU as the activation function. 
The model is trained on a single NVIDIA Tesla V100 GPU with 32GB memory.
The code will be open sourced on https://github.com/Anoise/Act-Now.

\subsection{Baselines } We consider a suite of baselines from continual learning, time series forecasting, and online learning, including OnlineTCN \cite{zinkevich2003online}, Experience Replay (ER) \cite{lin1992self,chaudhry2019tiny}, MIR \cite{aljundi2019online}, DER++ \cite{buzzega2020dark} and FSNet \cite{pham2023learning}. 
Furthermore, many advanced models are adopted as baselines for vafiey the Act-Now framework, including PSLD \cite{LIANG2024112622}, FreTS \cite{yi2024frequency}, FourierGNN \cite{yi2024fouriergnn}, Periodformer \cite{liang2023does}, DLinear \cite{zeng2023transformers}, FEDformer \cite{zhou2022fedformer}, Autoformer \cite{wu2021autoformer}, Informer \cite{Zhou2021Informer}.
All models use the same settings in the Act-Now framework. For example, the RSS algorithm is used in the data sampling process, and the input length is 36.
The models used in the experiments are evaluated over a wide range of prediction lengths to compare performance on different future horizons: 4, 6, and 8 for the C2TM dataset since the length of this dataset is short, and 24, 48, and 72 for others. The experimental settings and backbone are the same for all methods. 

\begin{figure*}
    \centerline{\includegraphics[width=0.9\textwidth]{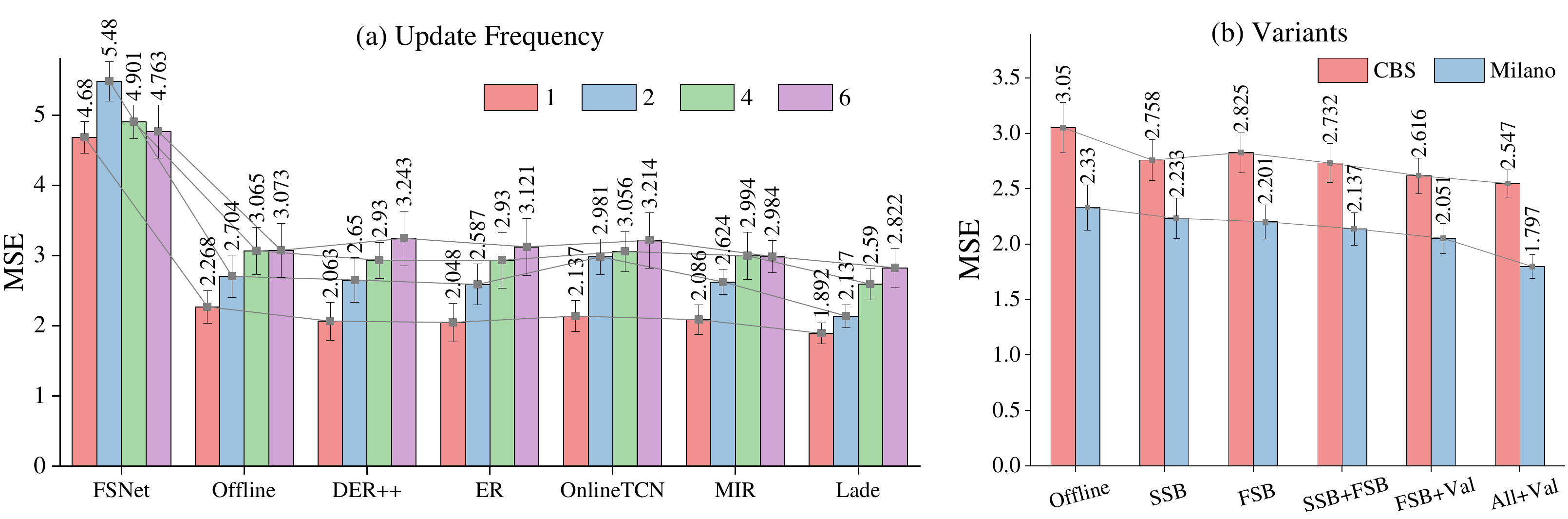}}
    \caption{(a) Ablation studies on the update frequency of streaming data using the Milano datasets. (b) Ablation studies on the components of Lade using CBS and Milano datasets. All+Val indicates the use of all components and updates on the validation set. All methods are performed at least three times.
    }
    \label{fig_freq}
\end{figure*}

\subsection{Main Results}

The online forecasting results for large-scale streaming datasets are summarized in Table \ref{tb_on_main}, which presents the cumulative performance of different baselines in terms of mean-squared errors (MSE) and mean-absolute errors (MAE).
It is evident that Lade consistently achieves state-of-the-art (SOTA) performance across all datasets and prediction length settings. Compared to the advanced ER and MIR, the proposed Lade yields an overall {\bf 28.4\%} and {\bf 19.5\%} relative cumulative MSE reduction on the CBS and Milano datasets, respectively. 
Specifically, for the input-36-predict-24 setting in CBS, Lade gives {\bf 14.5\%} (2.979$\rightarrow$2.547, compared to ER) and {\bf 13.9\%} (2.958$\rightarrow$2.547, compared to MIR) MSE reduction.
For the input-36-predict-24 setting on the Milano dataset, Lade gives {\bf 10.6\%} (1.988$\rightarrow$1.797, compared to ER) and {\bf 8.9\%} (1.973$\rightarrow$1.797, compared to MIR) MSE reduction.
For the C2TM dataset, which exhibits significant data fluctuations and a high noise level, Lade performs worse than the offline model on the input-8-output-4 setup but remains superior in other configurations.

\textit{Notice}: When the information leakage is removed, many online learning methods no longer work properly.
The degradation in performance stems from their reliance on future or unseen data during training. Information leakage often provides these methods with implicit access to patterns or trends that are not genuinely available in a true streaming scenario, allowing them to optimize their models unrealistically. Without leakage, these methods must adapt to data in a strictly causal manner, relying solely on historical and present data for updates. This restriction exposes their inability to effectively handle challenges like delayed feedback and concept drift in a dynamic setting.

\subsection{Ablation Studies on the Update Frequency}

The update frequency determines the availability of labels over time and the timing of model updates, which is crucial for online forecasting methods.
We conduct ablation studies on the update frequency of streaming data using the Milano dataset. The experimental results are shown in Fig. \ref{fig_freq}(a). 
It can be observed that as the update frequency of streaming data increases, the model's performance gradually improves. This is because the model needs to skip $D_{freq}$ time steps for prediction, which delays exposure to the true labels compared to the one-step-ahead rolling prediction. As a result, model updates occur more slowly, leading to performance degradation.
Note that FSNet is no longer used as a reference in this scenario because its fast and slow learning mechanisms become ineffective after removing the information leakage.

In addition, we observe that Lade consistently outperforms other online learning methods, regardless of changes in the update frequency. 
It is suggested that Lade effectively balances responsiveness to new data with stability in its learning process, enabling it to maintain high performance even when updates occur at irregular intervals.

\begin{table}[!t]
\centering
\caption{Ablation studies on the number of subgraph partitions of RSS.}
\label{tb_part}
\resizebox{\columnwidth}{!}
{
    \begin{tabular}{cc| cc|cc|cc}
    \toprule
    \multicolumn{2}{c|}{Partitions}       & \multicolumn{2}{c|}{10} & \multicolumn{2}{c|}{20} & \multicolumn{2}{c}{25} \\ \midrule
    Methods                    & Lengths & MSE        & MAE       & MSE        & MAE       & MSE        & MAE       \\  \midrule
    \multirow{2}{*}{Offline}   & 24      & 2.268      & 0.706     & 2.330      & 0.738     & 2.357      & 0.729     \\
                               & 48      & 2.837      & 0.754     & 3.191      & 0.792     & 3.162      & 0.793     \\ \midrule
    \multirow{2}{*}{DER++}     & 24      & 2.063      & 0.628     & 2.139      & 0.625     & 2.386      & 0.610     \\
                               & 48      & 2.548      & 0.692     & 2.543      & 0.664     & 2.514      & 0.658     \\ \midrule
    \multirow{2}{*}{ER}        & 24      & 2.048      & 0.618     & 1.988      & 0.593     & 2.038      & 0.550     \\
                               & 48      & 2.529      & 0.681     & 2.473      & 0.642     & 2.445      & 0.635     \\ \midrule
    \multirow{2}{*}{FSNet}     & 24      & 4.680      & 0.927     & 4.001      & 0.896     & 3.936      & 0.878     \\
                               & 48      & 5.187      & 0.986     & 4.723      & 0.935     & 4.727      & 0.930     \\ \midrule
    \multirow{2}{*}{OnlineTCN} & 24      & 2.137      & 0.675     & 2.322      & 0.713     & 2.312      & 0.708     \\
                               & 48      & 2.831      & 0.747     & 3.190      & 0.779     & 3.134      & 0.777     \\ \midrule
    \multirow{2}{*}{MIR}       & 24      & 2.086      & 0.648     & 1.973      & 0.589     & 1.957      & 0.581     \\
                               & 48      & 2.424      & 0.644     & 2.371      & 0.621     & 2.347      & 0.614     \\ \midrule
    \multirow{2}{*}{Lade}      & 24      &\bf 1.802   &\bf 0.467  &\bf 1.803   &\bf 0.446  &\bf 1.797   &\bf 0.439   \\
                               & 48      &\bf 2.210   &\bf 0.482  &\bf 2.214   &\bf 0.488  &\bf 2.196   &\bf 0.481   \\ \bottomrule
    \end{tabular}
}
\end{table}

\subsection{Ablation Studies on the Components of Lade}

To validate the effectiveness of Lade, we conduct comprehensive ablation studies encompassing both component replacement and component removal experiments, as shown in Fig. \ref{fig_freq}(b).
It becomes evident that the model's average performance is superior when employing SSB, FSB, and Val (online updates on the validation set).
E.g., on the Milano dataset, forecast error is reduced by {\bf 4.2\%} (2.33$\rightarrow$2.233).
Moreover, the introduction of FSB, further enhances the model's performance, e.g., forecast error on the Milano dataset is reduced by {\bf 1.4\%} (2.233$\rightarrow$2.201).
Afterward, incorporating Val holds the potential to improve predictive performance again, e.g., on the Milano dataset, forecast error is continuous reduced by {\bf 6.8\%} (2.201$\rightarrow$2.051).
Finally, using SSB+FFB+Val, the forecast error was further reduced by {\bf 12.4\%} (2.051$\rightarrow$1.797), demonstrating the effectiveness of each component.
In summary, integrating the components has the potential to significantly boost the model's performance across the board.

\subsection{Ablation Studies on Subgraph Partitioning}

We conducted ablation experiments with different subgraph sizes on the Milano datasets.
Firstly, we initially employ the RSS algorithm to perform data sampling. Then, in RSS algorithm, we random decompose the entire large graph into 10, 20 and 25 subgraphs at each sampling. 
As shown in Table \ref{tb_part}, the experimental findings indicate that the model's performance is stable across different subgraph sizes, reinforcing the robustness of the subgraph partitioning. This robustness indicates that the RSS algorithm provides comprehensive coverage of the original large-scale graph. It serves as an unbiased estimator of the true aggregated features derived from the entire graph, irrespective of the partitioning scheme, enabling the model to generalize effectively across different graph sizes.
Such stability is crucial for scalability, as it assures that the algorithm can handle large and complex graphs without significant performance degradation, making it suitable for real-world applications where the graph size may fluctuate or where different granularities of decomposition are required.

\subsection{Versatility of Act-Now Framework}

We integrate other advanced time series forecasting models into the Act-Now framework for online forecasting. The experimental results are shown in Table \ref{tb_ts_models}. Our observations indicate that the Act-Now framework significantly enhances the predictive performance of these advanced models.
For example, after adopting the Act-Now framework, the performance of Informer improved by 56\% and 41\% for prediction lengths of 24 and 48, respectively. Moreover, other advanced models also achieved at least a 2\% performance improvement.
The conducted experiments suggest that Act-Now framework can serve as a versatile framework, amenable to the integration of novel modules, thereby facilitating the enhancement of performance in the domain of TS forecasting.

It is worth noting that Lade achieved state-of-the-art performance in both online and offline learning scenarios. This demonstrates that the statistical and normalization flow design adopted by Lade effectively addresses the issue of concept drift in streaming data.

\subsection{Comparative Analysis} 

Several groundbreaking models have demonstrated competitive performance on specific datasets under certain conditions. For example, Informer shows relatively poor results on large-scale datasets when tested across different horizon settings. This under performance can be attributed to the substantial concept drift present in each variate of these datasets, which renders the KL-divergence-based ProbSparse Attention, employed by Informer, ineffective. On the other hand, linear-based models like DLinear have shown promising results on large-scale datasets across varying horizon settings, while frequency-based models such as FreTS and FourierGNN have also produced favorable outcomes. This phenomenon arises from a twofold interplay of factors. Firstly, large-scale streaming data exhibits stronger concept drift, including increased non-periodicity and noise, which directly affects model generalization, particularly on the C2TM dataset. Secondly, other models tend to overfit non-stationary, large-scale data characterized by aperiodic fluctuations. Notably, Lade effectively addresses the challenges posed by concept drift in large-scale forecasting, significantly improving performance. Specifically, on large-scale datasets with numerous variables, Lade excels by employing label decomposition to extract multiple easy-to-learn components. These components are learned progressively at shallow layers and then combined at deeper layers, offering an effective solution to the concept drift challenge in large-scale streaming data.

\begin{table}[!t]
\centering
\caption{The performance improvement (IMP) of the IL framework over other time series forecasting models on the Milano dataset.}
\label{tb_ts_models}
\resizebox{\columnwidth}{!}
{
    \begin{tabular}{c|ccc|ccc}
    \toprule
    Lengths      & \multicolumn{3}{c|}{24}  & \multicolumn{3}{c}{48}  \\ \midrule
    Methods      & Offline & Online & IMP  & Offline & Online & IMP  \\ \midrule
    Informer     & 12.028  & 5.328  & 56\% & 19.721  & 11.585 & 41\% \\
    Autoformer   & 3.264   & 2.395  & 27\% & 5.514   & 3.680  & 33\% \\
    FEDformer    & 3.222   & 2.457  & 24\% & 5.219   & 3.597  & 31\% \\
    Periodformer & 3.211   & 2.612  & 19\% & 4.679   & 3.294  & 30\% \\
    Dlinear      & 2.133   & 2.074  & 3\%  & 2.650   & 2.524  & 5\%  \\
    FreTS        & 2.000   & 1.934  & 3\%  & 2.629   & 2.285  & 13\% \\
    FourierGNN   & 1.945   & 1.912  & 2\%  & 2.682   & 2.435  & 9\%  \\
    PSLD         & 1.912   & 1.852  & 3\%  & 2.625   & 2.369  & 10\% \\
    Lade         &\bf 1.841&\bf 1.802& 2\% &\bf 2.607&\bf 2.196& 16\% \\ \bottomrule
    \end{tabular}
}
\end{table}

\subsection{Efficiency}

As shown in Table \ref{tb_times}, the proposed Lade framework achieves superior performance, offering significantly faster inference compared to most competing models. While the purely linear DLinear model attains the fastest training and inference speeds, it does so at the expense of accuracy. In contrast, Transformer-based models exhibit the highest computational complexity, leading to significantly slower training and inference times as well as reduced performance.
On the contrary, Lade strikes an effective balance between speed (training and inference) and performance (lowest MSE). This makes Lade particularly well-suited for scenarios that demand both high predictive performance and moderate computational resource usage.

\begin{table}[!ht]
    \centering
    \caption{Comparison of model's training time (Seconds/Epoch), inference time (Seconds/Epoch), parameters, FLOPs and MSE.}
    \label{tb_times}
    \resizebox{\columnwidth}{!}{
    \begin{tabular}{cccccc}
      \toprule
        Models       & Training (S) & Inference (S) & Params (M) & Flops (G) & MSE    \\ \hline
        Informer     & 36.75             & 12.81              & 12.79      & 2.13      & 20.768 \\
        Autoformer   & 37.6              & 15.18              & 12.63      & 2.43      & 2.296  \\
        FEDformer    & 80.89             & 31.28              & 12.63      & 2.43      & 2.336  \\
        Periodformer & 26.55             & 13.79              & 8.95       & 1.73      & 1.841  \\
        DLinear      & 2.51              & 0.95               & 1.8        & 0.0367    & 1.598  \\
        PSLD         & 6.86              & 2.38               & 2.0        & 0.232     & 0.633  \\ 
        Lade         & 5.75              & 2.17               & 1.93        & 0.210    & 0.508  \\ \bottomrule
    \end{tabular}
  }
  \end{table}

\subsection{Visualization the Learning Process of Lade}

\begin{figure*}
  \centerline{\includegraphics[width=\textwidth]{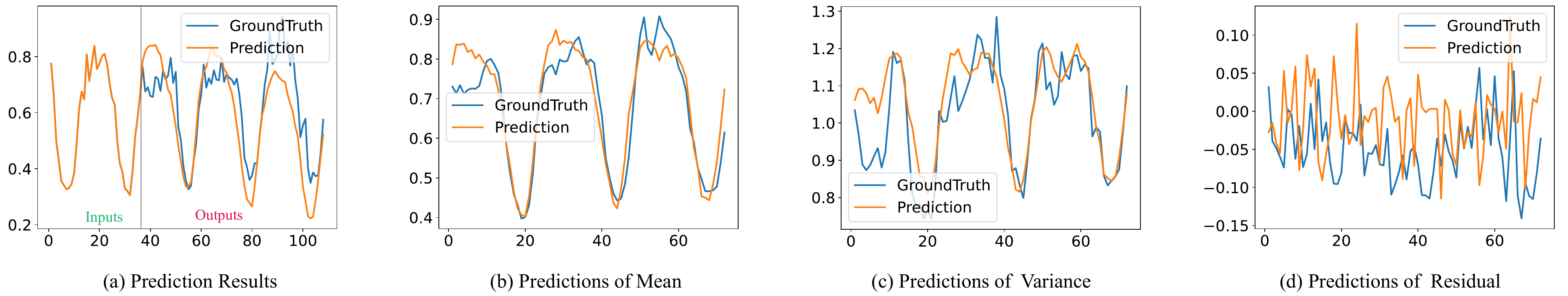}}
  \caption{Visualization of Lade prediction results utilizing MVD. Prediction cases from the Milano dataset under the input-36-predict-72 setting.}
  \label{fig_vis_mvd} %  A lower MSE indicates a better performance. 
\end{figure*}

The intrinsic feature of Lade lies in its alignment of each component’s output with the structure of the label’s decompositions. This alignment facilitates both the decomposition process and the visualization of the model's learning dynamics. As illustrated in Fig. \ref{fig_vis_mvd}, the outputs of Lade's components are visualized: Fig. \ref{fig_vis_mvd}(a) presents the predictions of MVD, Fig. \ref{fig_vis_mvd}(b) depicts the predictions of the mean, Fig. \ref{fig_vis_mvd}(c) illustrates the variance predictions, and Fig. \ref{fig_vis_mvd}(d) displays the residual predictions. These visualizations demonstrate that each component effectively identifies and incorporates meaningful patterns within the series. Decomposing $Y$ into its mean and variance provides valuable insights into the underlying data structure. The mean represents the central tendency, while the variance captures the data's spread or dispersion. For example, in Fig. \ref{fig_vis_mvd}, despite differences in series amplitudes, the mean and variance components are accurately fitted to their ground-truth values. Specifically, as shown in Figs. \ref{fig_vis_mvd}(a) and \ref{fig_vis_mvd}(b), for non-stationary series with varying means and variances, each block learns salient patterns. The mean component effectively captures trends and seasonality, while the variance component accounts for temporal changes in variability.
Additionally, predicting the variance enables the quantification of prediction uncertainty, paving the way for more robust and reliable forecasting.

\subsection{Visualization of Forecasting Results}

\begin{figure*}
  \centerline{\includegraphics[width=\textwidth]{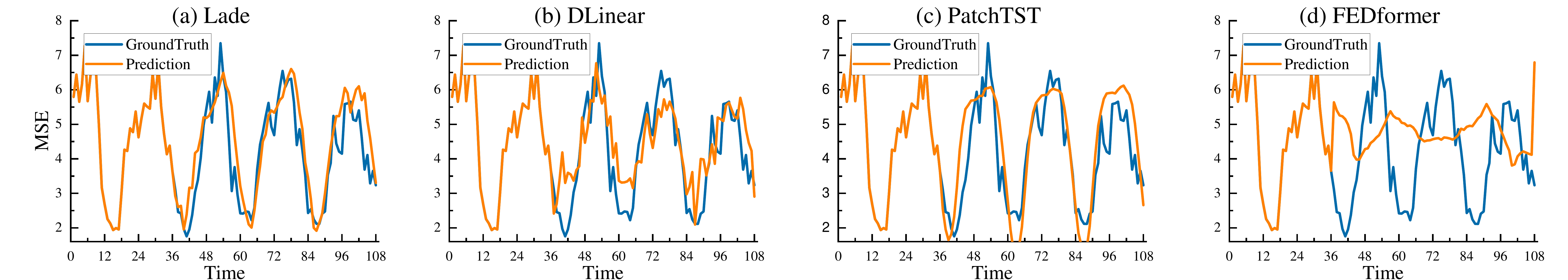}}
  \caption{Visualization of the prediction results across multiple models. Prediction cases from the Milano dataset under the input-36-predict-72 setting.}
  \label{fig_vis_internet}
\end{figure*}

For clarity and comparison among different models, we present supplementary prediction showcases in Fig. \ref{fig_vis_internet}. Visualization of different models for qualitative comparisons. Prediction cases from Milano dataset. This showcases correspond to predictions made by the following models: DLinear \cite{zeng2023transformers}, PatchTST \cite{nietime2023ICLR}, and FEDformer \cite{zhou2022fedformer}. Among the various models considered, the proposed Lade stands out for its ability to predict future series variations with exceptional precision, demonstrating superior performance.

\section{Conclusion}
\label{sec_con}

This paper identifies several key issues in existing online forecasting methods, such as information leakage, performance degradation due to online parameter updates, the exclusion of a validation set, and limitations in GPU devices for handling large-scale streaming data. To address these challenges, we propose a novel online learning framework (Act-Now), which enhances online forecasting for large-scale streaming data. The Act-Now framework includes several innovative components. 
First, we introduce the Random Subgraph Sampling (RSS) algorithm, which enables efficient model training by decomposing large graphs into manageable subgraphs. Then, we design two buffers—Fast Stream Buffer (FSB) and Slow Stream Buffer (SSB)—for online model updates. FSB enables immediate updates with consistent pseudo labels and partial labels, removing the issue of information leakage, while SSB performs parallel updates using complete labels from earlier time steps. 
Next, to combat concept drift, we propose the Label Decomposition (Lade) model, which forecasts both statistical variations and normalized future flows. Additionally, we perform online updates on the validation set to ensure continuous consistency in model learning over time. Our extensive experimental results demonstrate that the Act-Now framework significantly improves performance on large-scale streaming data, yielding average improvements of 28.4\% and 19.5\%, respectively, over existing methods.

\section*{Appendix}
\label{sec_app}

\subsection{Algorithm of RSS}
\label{ssec_rss_alg}

The pseudocode implementation of RSS in Algorithm~\ref{alg_rss}.

\begin{algorithm}[!ht]
  \caption{Python Pseudocode Implementation of RSS}
  \label{alg_rss}
  \begin{algorithmic}[1]
    \REQUIRE graph $G=(V,E)$, the number of nodes $N_{node}$, the number of subgraphs $N_{part}$, the length of the data $L_{data}$, input length $L_{in}$, output length $L_{out}$, and sampling frequency $D_{freq}$.
    \STATE  $N_{sub} \gets N_{node}/N_{part}$
    \STATE  $T \gets L_{data} - L_{in} - L_{out}$
    \FOR{$t$ in $T\times N_{part}$} 
        \IF{Training and Random Sampling} 
            \STATE  $I_{index} \gets  $\textbf{Randint}(0, $N_{node}$, $N_{sub}$)
            \STATE  $t \gets t \% T $
            \STATE $X \gets V[t:t+L_{in}, I_{index}]$
            \STATE $Y \gets V[t+L_{in}:t+L_{in}+L_{out}, I_{index}]$
            \IF{Using Edge}
                \STATE  $E_{row} \gets E[I_{index}]$
                \STATE  $E_{sub} \gets E_{row}[:,I_{index}]$
                \RETURN $X, Y, E_{sub}, t$
            \ENDIF
            \RETURN $X, Y, t$
        \ELSE
            \STATE  $I_{index} \gets t / (T\times D_{freq}) $
            \STATE  $N_{begin} \gets I_{index} \times N_{sub} $
            \STATE  $N_{end} \gets (I_{index}+1) \times N_{sub} $
            \STATE  $X \gets V[t:t+L_{in}, N_{begin}:N_{end}]$
            \STATE $Y \gets V[t+L_{in}:t+L_{in}+L_{out}, N_{begin}:N_{end}]$
            \IF{Using Edge}
                \STATE $I_{index} = \textbf{Arange}(N_{begiin}, N_{end})$
                \STATE  $E_{row} \gets E[I_{index}]$
                \STATE  $E_{sub} \gets E_{row}[:,I_{index}]$
                \RETURN $X, Y, E_{sub}, t$
            \ENDIF
            \RETURN $X, Y, t$
       \ENDIF
    \ENDFOR
  \end{algorithmic}
\end{algorithm}

\subsection{Proof of Theorem \ref{th1}}
\label{ssec_prf_th1}

Now we give proof of the representation ability of RSS.

\begin{proof}
Let $A(v)$ denote the true aggregated feature for node $v$ using the entire graph:
\begin{equation}
A(v) = \sum_{u \in N(v)} \frac{1}{C_{vu}} Wh_u,
\end{equation}
where $N(v)$ is the set of neighbors of $v$, $h_u$ is the feature of node $u$, 
$C_{vu}$ is a normalization constant, $W$ is the weight matrix.
For the sampled subgraph $G'$, the aggregated feature $A'(v)$ is given by:
\begin{equation}
  A'(v) = \sum_{u \in N(v) \cup V'} \frac{1}{C_{vu}P(u)} Wh_u,
\end{equation}
where $P(u)$ is the probability of node $u$ being sampled.
To show that $A'(v)$ is an unbiased estimator of $A(v)$, we compute the expectation of $A'(v)$:
\begin{equation}
  \mathbb{E}[A'(v)] = \mathbb{E}\left[ \sum_{u \in N(v) \cup V'} \frac{1}{C_{vu}P(u)} Wh_u \right].
\end{equation}
Since nodes are sampled independently, we can write:
\begin{equation}
  \mathbb{E}[A'(v)] = \mathbb{E}\left[ \sum_{u \in N(v)} \frac{\mathbb{I}(u\in V')}{C_{vu}P(u)} Wh_u \right],
\end{equation}
where $\mathbb{I}(u\in V')$ is an indicator function that is $1$ if 
$u$ is sampled and $0$ otherwise. The expectation of the indicator function is simply the probability of sampling $u$: $\mathbb{E}[\mathbb{I}(u\in V')]=P(u)$. Thus, 
\begin{align}
  \mathbb{E}[A'(v)] &= \sum_{u \in N(v)} \mathbb{E}\left[ \frac{\mathbb{I}(u\in V')}{C_{vu}P(u)} Wh_u \right] \notag \\
  &= \sum_{u \in N(v)} \frac{P(u)}{C_{vu}P(u)} Wh_u \notag \\
  &= \sum_{u \in N(v)} \frac{1}{C_{vu}} Wh_u. 
\end{align}
This shows that the expected aggregated feature using RSS is an unbiased estimator of the true aggregated feature using the entire graph.
\end{proof}

\bibliography{citation}

\begin{thebibliography}{10}
\providecommand{\url}[1]{#1}
\csname url@samestyle\endcsname
\providecommand{\newblock}{\relax}
\providecommand{\bibinfo}[2]{#2}
\providecommand{\BIBentrySTDinterwordspacing}{\spaceskip=0pt\relax}
\providecommand{\BIBentryALTinterwordstretchfactor}{4}
\providecommand{\BIBentryALTinterwordspacing}{\spaceskip=\fontdimen2\font plus
\BIBentryALTinterwordstretchfactor\fontdimen3\font minus \fontdimen4\font\relax}
\providecommand{\BIBforeignlanguage}[2]{{%
\expandafter\ifx\csname l@#1\endcsname\relax
\typeout{** WARNING: IEEEtran.bst: No hyphenation pattern has been}%
\typeout{** loaded for the language `#1'. Using the pattern for}%
\typeout{** the default language instead.}%
\else
\language=\csname l@#1\endcsname
\fi
#2}}
\providecommand{\BIBdecl}{\relax}
\BIBdecl

\bibitem{narayanan2021variegated}
A.~Narayanan, X.~Zhang, R.~Zhu, A.~Hassan, S.~Jin, X.~Zhu, X.~Zhang, D.~Rybkin, Z.~Yang, Z.~M. Mao \emph{et~al.}, ``A variegated look at 5g in the wild: performance, power, and qoe implications,'' in \emph{Proceedings of the 2021 ACM SIGCOMM 2021 Conference}, 2021, pp. 610--625.

\bibitem{asghar2022evolution}
M.~Z. Asghar, S.~A. Memon, and J.~H{\"a}m{\"a}l{\"a}inen, ``Evolution of wireless communication to 6g: Potential applications and research directions,'' \emph{Sustainability}, vol.~14, no.~10, p. 6356, 2022.

\bibitem{niu2010cell}
Z.~Niu, Y.~Wu, J.~Gong, and Z.~Yang, ``Cell zooming for cost-efficient green cellular networks,'' \emph{IEEE communications magazine}, vol.~48, no.~11, pp. 74--79, 2010.

\bibitem{kato2016deep}
N.~Kato, Z.~M. Fadlullah, B.~Mao, F.~Tang, O.~Akashi, T.~Inoue, and K.~Mizutani, ``The deep learning vision for heterogeneous network traffic control: Proposal, challenges, and future perspective,'' \emph{IEEE wireless communications}, vol.~24, no.~3, pp. 146--153, 2016.

\bibitem{DeepCog2019}
D.~Bega, M.~Gramaglia, M.~Fiore, A.~Banchs, and X.~Costa-Pérez, ``Deepcog: Optimizing resource provisioning in network slicing with ai-based capacity forecasting,'' \emph{IEEE Journal on Selected Areas in Communications}, vol.~38, no.~2, pp. 361--376, 2020.

\bibitem{li2017intelligent}
R.~Li, Z.~Zhao, X.~Zhou, G.~Ding, Y.~Chen, Z.~Wang, and H.~Zhang, ``Intelligent 5g: When cellular networks meet artificial intelligence,'' \emph{IEEE Wireless communications}, vol.~24, no.~5, pp. 175--183, 2017.

\bibitem{anava2013online}
O.~Anava, E.~Hazan, S.~Mannor, and O.~Shamir, ``Online learning for time series prediction,'' in \emph{Conference on learning theory}.\hskip 1em plus 0.5em minus 0.4em\relax PMLR, 2013, pp. 172--184.

\bibitem{liu2016online}
C.~Liu, S.~C. Hoi, P.~Zhao, and J.~Sun, ``Online arima algorithms for time series prediction,'' in \emph{Thirtieth AAAI conference on artificial intelligence}, 2016.

\bibitem{smola2003laplace}
A.~J. Smola, V.~Vishwanathan, and E.~Eskin, ``Laplace propagation.'' in \emph{NIPS}.\hskip 1em plus 0.5em minus 0.4em\relax Citeseer, 2003, pp. 441--448.

\bibitem{gultekin2018online}
S.~Gultekin and J.~Paisley, ``Online forecasting matrix factorization,'' \emph{IEEE Transactions on Signal Processing}, vol.~67, no.~5, pp. 1223--1236, 2018.

\bibitem{kurle2019continual}
R.~Kurle, B.~Cseke, A.~Klushyn, P.~Van Der~Smagt, and S.~G{\"u}nnemann, ``Continual learning with bayesian neural networks for non-stationary data,'' in \emph{International Conference on Learning Representations}, 2019.

\bibitem{aydore2019dynamic}
S.~Aydore, T.~Zhu, and D.~P. Foster, ``Dynamic local regret for non-convex online forecasting,'' \emph{Advances in Neural Information Processing Systems}, vol.~32, pp. 7982--7991, 2019.

\bibitem{woo2022cost}
G.~Woo, C.~Liu, D.~Sahoo, A.~Kumar, and S.~Hoi, ``Cost: Contrastive learning of disentangled seasonal-trend representations for time series forecasting,'' \emph{arXiv preprint arXiv:2202.01575}, 2022.

\bibitem{zhang2023onenet}
Y.-F. Zhang, Q.~Wen, X.~Wang, W.~Chen, L.~Sun, Z.~Zhang, L.~Wang, R.~Jin, and T.~Tan, ``Onenet: Enhancing time series forecasting models under concept drift by online ensembling,'' \emph{arXiv preprint arXiv:2309.12659}, 2023.

\bibitem{pham2023learning}
Q.~Pham, C.~Liu, D.~Sahoo, and S.~Hoi, ``Learning fast and slow for online time series forecasting,'' in \emph{The Eleventh International Conference on Learning Representations}, 2023.

\bibitem{taylor2018forecasting}
S.~J. Taylor and B.~Letham, ``Forecasting at scale,'' \emph{The American Statistician}, vol.~72, no.~1, pp. 37--45, 2018.

\bibitem{oreshkin2019n}
B.~N. Oreshkin, D.~Carpov, N.~Chapados, and Y.~Bengio, ``N-beats: Neural basis expansion analysis for interpretable time series forecasting,'' in \emph{International Conference on Learning Representations}, 2019.

\bibitem{sen2019think}
R.~Sen, H.-F. Yu, and I.~S. Dhillon, ``Think globally, act locally: A deep neural network approach to high-dimensional time series forecasting,'' \emph{Advances in neural information processing systems}, vol.~32, 2019.

\bibitem{wu2021autoformer}
H.~Wu, J.~Xu, J.~Wang, and M.~Long, ``Autoformer: Decomposition transformers with auto-correlation for long-term series forecasting,'' in \emph{Advances in Neural Information Processing Systems (NeurIPS)}, vol.~34, Virtual Conference, 2021, pp. 22\,419--22\,430.

\bibitem{zhou2022fedformer}
T.~Zhou, Z.~Ma, Q.~Wen, X.~Wang, L.~Sun, and R.~Jin, ``{FEDformer}: Frequency enhanced decomposed transformer for long-term series forecasting,'' in \emph{Proceedings of the 39th International Conference on Machine Learning (ICML)}, vol. 162, Baltimore, Maryland, 2022, pp. 27\,268--27\,286.

\bibitem{liu2022non}
Y.~Liu, H.~Wu, J.~Wang, and M.~Long, ``Non-stationary transformers: Exploring the stationarity in time series forecasting,'' \emph{Advances in Neural Information Processing Systems}, vol.~35, pp. 9881--9893, 2022.

\bibitem{tsymbal2004problem}
A.~Tsymbal, ``The problem of concept drift: definitions and related work,'' \emph{Computer Science Department, Trinity College Dublin}, vol. 106, no.~2, p.~58, 2004.

\bibitem{li2022ddg}
W.~Li, X.~Yang, W.~Liu, Y.~Xia, and J.~Bian, ``Ddg-da: Data distribution generation for predictable concept drift adaptation,'' in \emph{Proceedings of the AAAI Conference on Artificial Intelligence}, vol.~36, no.~4, 2022, pp. 4092--4100.

\bibitem{qin2022generalizing}
T.~Qin, S.~Wang, and H.~Li, ``Generalizing to evolving domains with latent structure-aware sequential autoencoder,'' in \emph{International Conference on Machine Learning}.\hskip 1em plus 0.5em minus 0.4em\relax PMLR, 2022, pp. 18\,062--18\,082.

\bibitem{sahoo2018online}
D.~Sahoo, Q.~Pham, J.~Lu, and S.~C. Hoi, ``Online deep learning: learning deep neural networks on the fly,'' in \emph{Proceedings of the 27th International Joint Conference on Artificial Intelligence}, 2018, pp. 2660--2666.

\bibitem{gama2014survey}
J.~Gama, I.~{\v{Z}}liobait{\.e}, A.~Bifet, M.~Pechenizkiy, and A.~Bouchachia, ``A survey on concept drift adaptation,'' \emph{ACM computing surveys (CSUR)}, vol.~46, no.~4, pp. 1--37, 2014.

\bibitem{huszar2017quadratic}
F.~Husz{\'a}r, ``On quadratic penalties in elastic weight consolidation,'' \emph{arXiv preprint arXiv:1712.03847}, 2017.

\bibitem{kirkpatrick2018reply}
J.~Kirkpatrick, R.~Pascanu, N.~Rabinowitz, J.~Veness, G.~Desjardins, A.~A. Rusu, K.~Milan, J.~Quan, T.~Ramalho, A.~Grabska-Barwinska \emph{et~al.}, ``Reply to husz{\'a}r: The elastic weight consolidation penalty is empirically valid,'' \emph{Proceedings of the National Academy of Sciences}, vol. 115, no.~11, pp. E2498--E2498, 2018.

\bibitem{lopez2017gradient}
D.~Lopez-Paz and M.~Ranzato, ``Gradient episodic memory for continual learning,'' \emph{Advances in neural information processing systems}, vol.~30, pp. 6467--6476, 2017.

\bibitem{grossberg1982does}
S.~Grossberg, ``How does a brain build a cognitive code?'' \emph{Studies of mind and brain}, pp. 1--52, 1982.

\bibitem{lin1992self}
L.-J. Lin, ``Self-improving reactive agents based on reinforcement learning, planning and teaching,'' \emph{Machine learning}, vol.~8, no. 3-4, pp. 293--321, 1992.

\bibitem{riemer2018learning}
M.~Riemer, I.~Cases, R.~Ajemian, M.~Liu, I.~Rish, Y.~Tu, and G.~Tesauro, ``Learning to learn without forgetting by maximizing transfer and minimizing interference,'' \emph{International Conference on Learning Representations (ICLR)}, 2019.

\bibitem{rolnick2019experience}
D.~Rolnick, A.~Ahuja, J.~Schwarz, T.~Lillicrap, and G.~Wayne, ``Experience replay for continual learning,'' in \emph{Advances in Neural Information Processing Systems}, 2019, pp. 348--358.

\bibitem{pham2021dualnet}
Q.~Pham, C.~Liu, and S.~Hoi, ``Dualnet: Continual learning, fast and slow,'' \emph{Advances in Neural Information Processing Systems}, vol.~34, 2021.

\bibitem{arani2021learning}
E.~Arani, F.~Sarfraz, and B.~Zonooz, ``Learning fast, learning slow: A general continual learning method based on complementary learning system,'' in \emph{International Conference on Learning Representations}, 2021.

\bibitem{cleveland1990stl}
R.~B. Cleveland, W.~S. Cleveland, J.~E. McRae, and I.~Terpenning, ``Stl: A seasonal-trend decomposition,'' \emph{J. Off. Stat}, vol.~6, no.~1, pp. 3--73, 1990.

\bibitem{de2011forecasting}
A.~M. De~Livera, R.~J. Hyndman, and R.~D. Snyder, ``Forecasting time series with complex seasonal patterns using exponential smoothing,'' \emph{Journal of the American statistical association}, vol. 106, no. 496, pp. 1513--1527, 2011.

\bibitem{cheng2015time}
C.~Cheng, A.~Sa-Ngasoongsong, O.~Beyca, T.~Le, H.~Yang, Z.~Kong, and S.~T. Bukkapatnam, ``Time series forecasting for nonlinear and non-stationary processes: a review and comparative study,'' \emph{Iie Transactions}, vol.~47, no.~10, pp. 1053--1071, 2015.

\bibitem{barlacchi2015multi}
G.~Barlacchi, M.~De~Nadai, R.~Larcher, A.~Casella, C.~Chitic, G.~Torrisi, F.~Antonelli, A.~Vespignani, A.~Pentland, and B.~Lepri, ``A multi-source dataset of urban life in the city of milan and the province of trentino,'' \emph{Scientific data}, vol.~2, no.~1, pp. 1--15, 2015.

\bibitem{Modeling15Chen}
Y.~J. Xiaming~Chen, S.~Qiang, W.~Hu, and K.~Jiang, ``Analyzing and modeling spatio-temporal dependence of cellular traffic at city scale,'' in \emph{Communications (ICC), 2015 IEEE International Conference on}, 2015.

\bibitem{kingma2014adam}
D.~P. Kingma and J.~Ba, ``Adam: A method for stochastic optimization,'' in \emph{International Conference on Learning Representations (ICLR)}, Santiago de Cuba, 2015.

\bibitem{zinkevich2003online}
M.~Zinkevich, ``Online convex programming and generalized infinitesimal gradient ascent,'' in \emph{Proceedings of the 20th international conference on machine learning (icml-03)}, 2003, pp. 928--936.

\bibitem{chaudhry2019tiny}
A.~Chaudhry, M.~Rohrbach, M.~Elhoseiny, T.~Ajanthan, P.~K. Dokania, P.~H. Torr, and M.~Ranzato, ``On tiny episodic memories in continual learning,'' \emph{arXiv preprint arXiv:1902.10486}, 2019.

\bibitem{aljundi2019online}
R.~Aljundi, E.~Belilovsky, T.~Tuytelaars, L.~Charlin, M.~Caccia, M.~Lin, and L.~Page-Caccia, ``Online continual learning with maximal interfered retrieval,'' \emph{Advances in Neural Information Processing Systems}, vol.~32, pp. 11\,849--11\,860, 2019.

\bibitem{buzzega2020dark}
P.~Buzzega, M.~Boschini, A.~Porrello, D.~Abati, and S.~Calderara, ``Dark experience for general continual learning: a strong, simple baseline,'' in \emph{34th Conference on Neural Information Processing Systems (NeurIPS 2020)}, 2020.

\bibitem{LIANG2024112622}
D.~Liang, H.~Zhang, D.~Yuan, and M.~Zhang, ``Progressive supervision via label decomposition: An long-term and large-scale wireless traffic forecasting method,'' \emph{Knowledge-Based Systems}, vol. 305, p. 112622, 2024.

\bibitem{yi2024frequency}
K.~Yi, Q.~Zhang, W.~Fan, S.~Wang, P.~Wang, H.~He, N.~An, D.~Lian, L.~Cao, and Z.~Niu, ``Frequency-domain mlps are more effective learners in time series forecasting,'' \emph{Advances in Neural Information Processing Systems}, vol.~36, 2024.

\bibitem{yi2024fouriergnn}
K.~Yi, Q.~Zhang, W.~Fan, H.~He, L.~Hu, P.~Wang, N.~An, L.~Cao, and Z.~Niu, ``Fouriergnn: Rethinking multivariate time series forecasting from a pure graph perspective,'' \emph{Advances in Neural Information Processing Systems}, vol.~36, 2024.

\bibitem{liang2023does}
D.~Liang, H.~Zhang, D.~Yuan, X.~Ma, D.~Li, and M.~Zhang, ``Does long-term series forecasting need complex attention and extra long inputs?'' \emph{arXiv preprint arXiv:2306.05035}, 2023.

\bibitem{zeng2023transformers}
A.~Zeng, M.~Chen, L.~Zhang, and Q.~Xu, ``Are transformers effective for time series forecasting?'' in \emph{Proceedings of the AAAI conference on artificial intelligence}, vol.~37, no.~9, 2023, pp. 11\,121--11\,128.

\bibitem{Zhou2021Informer}
H.~Zhou, S.~Zhang, J.~Peng, S.~Zhang, J.~Li, H.~Xiong, and W.~Zhang, ``Informer: Beyond efficient transformer for long sequence time-series forecasting,'' in \emph{Proceedings of the 35th AAAI Conference on Artificial Intelligence (AAAI)}, vol.~35, no.~12, Virtual Conference, 2021, pp. 11\,106--11\,115.

\bibitem{nietime2023ICLR}
Y.~Nie, N.~H. Nguyen, P.~Sinthong, and J.~Kalagnanam, ``A time series is worth 64 words: Long-term forecasting with transformers,'' in \emph{The Eleventh International Conference on Learning Representations}, 2023.

\end{thebibliography}

\end{document}